\documentclass[10pt,twocolumn,letterpaper]{article}

\usepackage{cvpr}
\usepackage{times}
\usepackage{epsfig}
\usepackage{graphicx}
\usepackage{amsmath}
\usepackage{amssymb}

\usepackage[pagebackref=true,breaklinks=true,letterpaper=true,colorlinks,bookmarks=false]{hyperref}

\cvprfinalcopy % *** Uncomment this line for the final submission

\def\cvprPaperID{864} % *** Enter the CVPR Paper ID here

\ifcvprfinal\fi
\begin{document}

%%%%%%%%% TITLE
\title{Batch-normalized Maxout Network in Network}

\author{Jia-Ren Chang\\
Department of Computer Science\\
National Chiao Tung University, \\Hsinchu, Taiwan\\
{\tt\small followwar.cs00g@nctu.edu.tw}
\and
Yong-Sheng Chen\\
Department of Computer Science\\
National Chiao Tung University, \\Hsinchu, Taiwan\\
{\tt\small yschen@cs.nctu.edu.tw}
\thanks{Corresponding author. This work was supported in part by the Taiwan Ministry of Science and Technology under Grant Number: MOST 103-2221-E-009 -131.}
}

\maketitle
%\thispagestyle{empty}

%%%%%%%%% ABSTRACT
\begin{abstract}
This paper reports a novel deep architecture referred to as Maxout network In Network (MIN), which can enhance model discriminability and facilitate the process of information abstraction within the receptive field. The proposed network adopts the framework of the recently developed Network In Network structure, which slides a universal approximator, multilayer perceptron (MLP) with rectifier units, to exact features. Instead of MLP, we employ maxout MLP to learn a variety of piecewise linear activation functions and to mediate the problem of vanishing gradients that can occur when using rectifier units. Moreover, batch normalization is applied to reduce the saturation of maxout units by pre-conditioning the model and dropout is applied to prevent overfitting. Finally, average pooling is used in all pooling layers to regularize maxout MLP in order to facilitate information abstraction in every receptive field while tolerating the change of object position. Because average pooling preserves all features in the local patch, the proposed MIN model can enforce the suppression of irrelevant information during training. Our experiments demonstrated the state-of-the-art classification performance when the MIN model was applied to MNIST, CIFAR-10, and CIFAR-100 datasets and comparable performance for SVHN dataset.
\end{abstract}

%%%%%%%%% BODY TEXT
\section{Introduction}
%	Deep learning has yielded significant advancements in image classification~\cite{corr:hinton2012improving} and speech recognition~\cite{maas2013rectifier}.
Deep convolutional neural networks (CNNs)~\cite{krizhevsky2012imagenet} have recently been applied to large image datasets, such as MNIST~\cite{lecun1998gradient}, CIFAR-10/100~\cite{krizhevsky2009learning}, SVHN~\cite{netzer2011reading}, and ImageNet~\cite{deng2009imagenet} for image classification~\cite{corr:hinton2012improving}. 
A deep CNN is able to learn basic filters automatically and combine them hierarchically to enable the description of latent concepts for pattern recognition.
In~\cite{zeiler2014visualizing}, Zeiler et al. illustrated how deep CNN organizes feature maps and the discrimination among classes. 

    Despite the advances that have been made in the development of this technology, many issues related to deep learning remain, including: (1) model discriminability and the robustness of learned features in early layers~\cite{zeiler2014visualizing}; (2) the vanishing gradients and saturation of activation units during training~\cite{glorot2010understanding}; and (3) limited training data, which may lead to overfitting~\cite{srivastava2014dropout}. 

	Because data are usually distributed on nonlinear manifolds, they are not separable by linear filters. For enhancing model discriminability, the Network In Network (NIN)~\cite{DBLP:journals/corr/LinCY13} model uses a sliding micro neural network, multilayer perceptron (MLP), to increase the nonlinearity of local patches in order to enable the abstraction of greater quantities of information within the receptive fields. Similarly, Deeply Supervised Nets (DSN)~\cite{lee2014deeply} provides companion objective functions to constrain hidden layers, such that robust features can be learned in the early layers of a deep CNN structure. 

The problem of vanishing gradients is essentially the shrinking of gradients backward through hidden layers. Some activation functions, such as sigmoid, are susceptible to vanishing gradients and saturation during the training of deep networks, due to the fact that higher hidden units become saturated at -1 or 1. Current solutions involve the use of rectified linear units (ReLU) to prevent vanishing gradients~\cite{krizhevsky2012imagenet,maas2013rectifier,nair2010rectified} because ReLU activates above 0 and its partial derivative is 1. Thus gradients flow through while ReLU activates. Unfortunately, ReLU has a potential disadvantage. The constant 0 will block the gradient flowing through inactivated ReLUs, such that some units may never activate. Recently, the maxout network~\cite{goodfellow2013maxout} provided a remedy to this problem. Even when maxout output is 0, this value is from a maxout hidden unit and this unit may be adjusted to become positive afterwards. Another issue involves changes of data distribution during the training of deep networks that are likely to saturate the activation function. This changed data distribution can move input data into the saturated regime of the activation function and slow down the training process. This phenomenon is referred to as internal covariate shift~\cite{shimodaira2000improving}. Ioffe et al.~\cite{ioffe2015batch} addressed this problem by applying batch normalization to the input of each layer. 

Dropout~\cite{srivastava2014dropout} and Dropconnect~\cite{wan2013regularization} techniques are widely used to regularize deep networks in order to prevent overfitting. The idea of the technique is to randomly drop units or connections to prevent units from co-adapting, which has been shown to improve classification accuracy in numerous studies~\cite{goodfellow2013maxout,lee2014deeply,DBLP:journals/corr/LinCY13,srivastava2014dropout}.

The previous  deep learning methods used max pooling to retain the most valuable features in the local patch. The max pooling mimics the spatial selective attention mechanism of human and attends to the important and discriminable areas of the input image. Lin et al.~\cite{DBLP:journals/corr/LinCY13} proposed a strategy referred to as global average pooling for the replacement of fully-connected layers to enable the summing of spatial information of feature maps, thereby making it more robust to spatial translation. This strategy is close to human visual process, in which retinotopic response can be predicted by linear spatial summation~\cite{hansen2004parametric}. The spatial summation enables the tolerance to the changes in object position and size, which is an essential characteristic to object recognition. In this work, we extend the global average pooling strategy to local spatial average pooling in order to aggregate local information from feature maps. 

However, the spatial average pooling keeps the irrelevant features together with the relevant ones and may deteriorate the classification performance.
This problem can be tackled by applying the maxout MLP to select the most prominent features from local patches before spatial average pooling. 
During the training process, therefore, the back-propagation enforces the maxout MLP to learn the most relevant features in every local patches.
From neuroscience perspective, there is a similar mechanism in visual system.
Within the same receptive field, the features of objects compete with each other in the object recognition process~\cite{desimone1995neural}.

In this study, we aimed to increase nonlinearity within local patches and alleviate the problem of vanishing gradients. Based on the NIN~\cite{DBLP:journals/corr/LinCY13} structure, we employ a maxout MLP for feature extraction and refer to the proposed model as Maxout network In Network (MIN).
The MIN model uses batch normalization to reduce saturation and uses dropout to prevent overfitting.
To increase the robustness to spatial translation of objects, furthermore, average pooling is applied in all pooling layers to aggregate the essential features obtained by maxout MLP.

%------------------------------------------------------------------------
\section{Design of Deep Architecture}
	This section presents previous works related to the proposed MIN structure, including NIN, Maxout Network, and batch normalization, followed by the design of the MIN architecture.

%-------------------------------------------------------------------------
\subsection{NIN}

	The NIN model~\cite{DBLP:journals/corr/LinCY13} uses the universal approximator MLP for the extraction of features from local patches.
Compared to CNN, MLP, wherein an ReLU is used as the activation function, enables the abstraction of information that is more representative for the latent concepts.
The NIN model introduced the $mlpconv$ layer which consists of a linear convolutional layer and a two-layer MLP. The calculation performed by the $mlpconv$ layer is as follows:

\[f^1_{i,j,n_1}={\rm max}\left({{{\mathbf w}}^1_{n_1}}^{{\rm T}}{{\mathbf x}}_{i,j}+b_{n_1},0\right)\ ,\] 
\[f^2_{i,j,n_2}={\rm max}\left({{{\mathbf w}}^2_{n_2}}^{{\rm T}}{{\mathbf f}}^1_{i,j}+b_{n_2},0\right)\ ,\]
\begin{equation}
f^3_{i,j,n_3}={\rm max}\left({{{\mathbf w}}^3_{n_3}}^{{\rm T}}{{\mathbf f}}^2_{i,j}+b_{n_3},0\right)\ ,
\end{equation}
where ${\rm (}i,j{\rm )}$ is the pixel index in the feature maps, ${{\mathbf x}}_{i,j}$ represents the input patch centered at location ${\rm (}i,j{\rm )}$, and $n_1,n_2,\ $ and $n_3$ are used to index the channels of the feature maps. From another perspective, the $mlpconv$ layer can be viewed as equivalent to a cascaded cross-channel parametric pooling layer on a convolutional layer. The cascaded cross-channel parametric pooling layer linearly combines feature maps and then passes through ReLUs, thereby allowing the cross-channel flow of information.

However, the constant 0 will block the gradients flowing through the inactivated ReLUs and these ReLUs will not be updated during the training process.
In this work, we adopted a similar universal approximator, maxout MLP, to overcome this problem.

%---------------------------------------------------
\subsection{Maxout Network}
Maxout MLP has previously been proven as a universal approximator~\cite{goodfellow2013maxout}, wherein a maxout unit is implemented by the following function:

\begin{equation}
f_{i,j,n}={\mathop{\max }_{m\in [1,k]} \left({{\mathbf w}}^{{\rm T}}_{n_m}{{\mathbf x}}_{i,j}+b_{n_m}\right)\ } ,
\end{equation}
where ${(i,j)}$ is the pixel index in the feature maps, ${{\mathbf x}}_{i,j}$ represents the input patch centered at location ${(i,j)}$, and \textit{n} is used to index the channels of the feature maps, $f_{i,j,n}$, which are constructed by taking the maximum across \textit{k} maxout hidden pieces. From another perspective, maxout unit is equivalent to a cross-channel max pooling layer on a convolutional layer. The cross-channel max pooling layer selects the maximal output to be fed into the next layer. The maxout unit is helpful to tackle the problem of vanishing gradients because the gradient is able to flow through every maxout unit.

Internal covariate shift is defined as changes in the distribution of network activations resulting from the updating of network parameters during training~\cite{ioffe2015batch}. When more weights and biases in a network are changed, internal covariate shift is severer.
A greater number of inputs move to the saturated regime of nonlinearity and thereby slow down the training process. Internal covariate shift is a serious problem for maxout MLP because it is \textit{k} times larger than a classical MLP. In this study, we applied batch normalization to reduce the effects of covariate shift.

%-----------------------------------------------------
\subsection{Batch Normalization}
	Batch normalization~\cite{ioffe2015batch} is used to independently normalize each channel toward zero mean and unit variance:
\begin{equation}
	 \hat{x}_{i,j,n}=\frac{x_{i,j,n}-\rm E\left[x_n\right]}{\sqrt{\rm Var\left[x_n\right]}} \,\, ,
\end{equation}
whereupon the normalized value undergoes scaling and shifting:
\begin{equation}
	 f_{i,j,n}=\gamma_n{\hat{x}}_{i,j,n}+\beta_n  \,\, .
\end{equation}
Here  \begin{math} x_{i,j} \end{math} stands for the value at location  \begin{math} (i,j)\end{math},  \begin{math} n\end{math} is used to index the channels of the feature map, and scaling and shifting parameters  \begin{math} {\gamma_n,\beta_n} \end{math} are new parameters that join in network training.

Batch normalization layer can be applied to a convolutional network immediately before the activation function, such as ReLU or maxout.
In this case, the nonlinearity units tends to produce activation with a stable distribution, which reduces saturation.
Normalization also exists in biological neural network, which is a canonical neural computation well-studied in neuroscience field~\cite{carandini2012normalization}.
This mechanism explains how primary visual neurons control the strength of input at which responses saturate.

%--------------------------------------------------------------------
\subsection{Proposed MIN Architecture}

The NIN~\cite{DBLP:journals/corr/LinCY13} has capability to abstract representative features within the receptive field and thereby achieve good results in image classification. As described in Section 2.1, NIN uses ReLU as the activation function in $mlpconv$ layer. In this study, we replaced the ReLU activation functions in the two-layer MLP in NIN with the maxout units to overcome the vanishing gradient problem commonly encountered when using ReLU. Furthermore, we applied batch normalization immediately after convolutional calculation to avoid the covariate shift problem caused by the changes of data distribution. Specifically, we removed the activation function of the convolutional layer, thereby rendering it a pure feature extractor. The architecture of the proposed MIN model is presented in Figure \ref{fig:MINfig}.  Feature maps in a MIN block are calculated as follows:

\begin{figure*}
\begin{center}
	\includegraphics*[width=6.8in, height=5.1in, keepaspectratio=false]{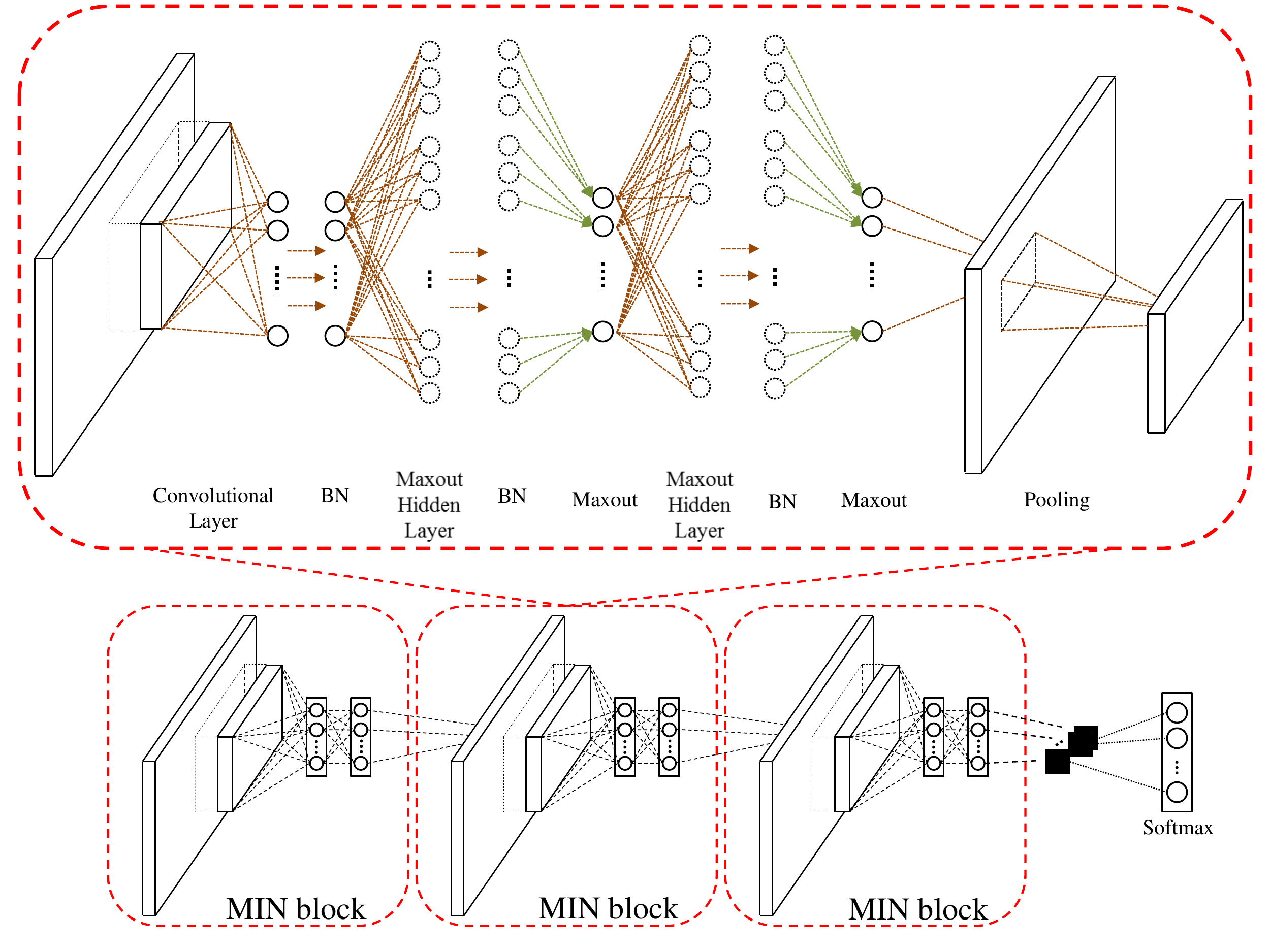}
\end{center}
   \caption{The architecture of the proposed MIN model.}
\label{fig:MINfig}
\end{figure*}

\[f^1_{i,j,n_1}={\rm BN}\left({{{\mathbf w}}^1_{n_1}}^{{\rm T}}{{\mathbf x}}_{i,j}+b^1_{n_j}\right)\ ,\] 
\[f^2_{{i,j,n}_2}={\mathop{\max }_{m\in [1,k_1]} \left({\rm BN}\left({{{\mathbf w}}^2_{n_m}}^{{\rm T}}{{\mathbf f}}^1_{i,j}+b^2_{n_m}\right)\right)\ }\ ,\] 
\begin{equation}
f^3_{i,j,n_3}={\mathop{\max }_{m\in [1,k_2]} \left({\rm BN}\left({{{\mathbf w}}^3_{n_m}}^{{\rm T}}{{\mathbf f}}^2_{i,j}+b^3_{n_m}\right)\right)\ }\ ,
\end{equation}
where ${\rm BN}\left(\cdot \right)$ denotes the batch normalization layer, $(i,j)$ is the pixel index in the feature map, ${{\mathbf x}}_{i,j}$ represents the input patch centered at location $(i,j)$, and \textit{n} is used to index the channels of the feature maps that are constructed by taking the maximum across \textit{k} maxout hidden pieces. Montufar et al. [18] demonstrated that the complexity of maxout networks increases with the number of maxout pieces or layers. By increasing the number of maxout pieces, the proposed model expands the ability to capture the latent concepts for various inputs.

	From another perspective, a MIN block is equivalent to a cascaded cross-channel parametric pooling layer and a cross-channel max pooling on a convolutional layer. The MIN block linearly combines feature maps and selects the combinations that are the most informational to be fed into the next layer. The MIN block reduces saturation by applying batch normalization and makes it possible to encode information on pathways or in the activation patterns of maxout pieces~\cite{wang2014maxout}. This makes it possible to enhance the discrimination capability of deep architectures.

%%-------------------
\section{Experiments}
In the following experiments, the proposed method was evaluated using four benchmark datasets: MNIST~\cite{lecun1998gradient}, CIFAR-10~\cite{krizhevsky2009learning}, CIFAR-100~\cite{krizhevsky2009learning}, and SVHN~\cite{netzer2011reading}.
The proposed model consists of three stacked MIN blocks followed by a softmax layer.
A MIN block includes a convolutional layer, a two-layer maxout MLP, and a spatial pooling layer. Dropout is applied between MIN blocks for regularization.
Table \ref{fig:para} details the parameter settings which, for the sake of a fair comparison, are the same as those used in NIN~\cite{DBLP:journals/corr/LinCY13}.
The network was implemented using the MatConvNet~\cite{vedaldi2014matconvnet} toolbox in the Matlab environment. We adopted the training procedure proposed by Goodfellow et al.~\cite{goodfellow2013maxout} to determine the hyper-parameters of the model, such as momentum, weight decay, and learning rate decay. 

\begin{table}
   \caption{Parameter settings of the proposed MIN architecture used in the experiments. The convolutional kernel is defined as (height) x (width) x (number of units). Below, we present the stride (st.), padding (pad) and batch normalization (BN) of the convolution kernel. In maxout MLP layers (MMLP), \textit{k} indicates the number of maxout pieces used in one maxout unit. A softmax layer is applied to the last layer in the model (not shown here). The top row lists the parameters used in CIFAR-10/100 and SVHN, whereas the bottom row lists those used in MNIST.}
\begin{center}
	\includegraphics*[width=2.725in, height=3.2in, keepaspectratio=false]{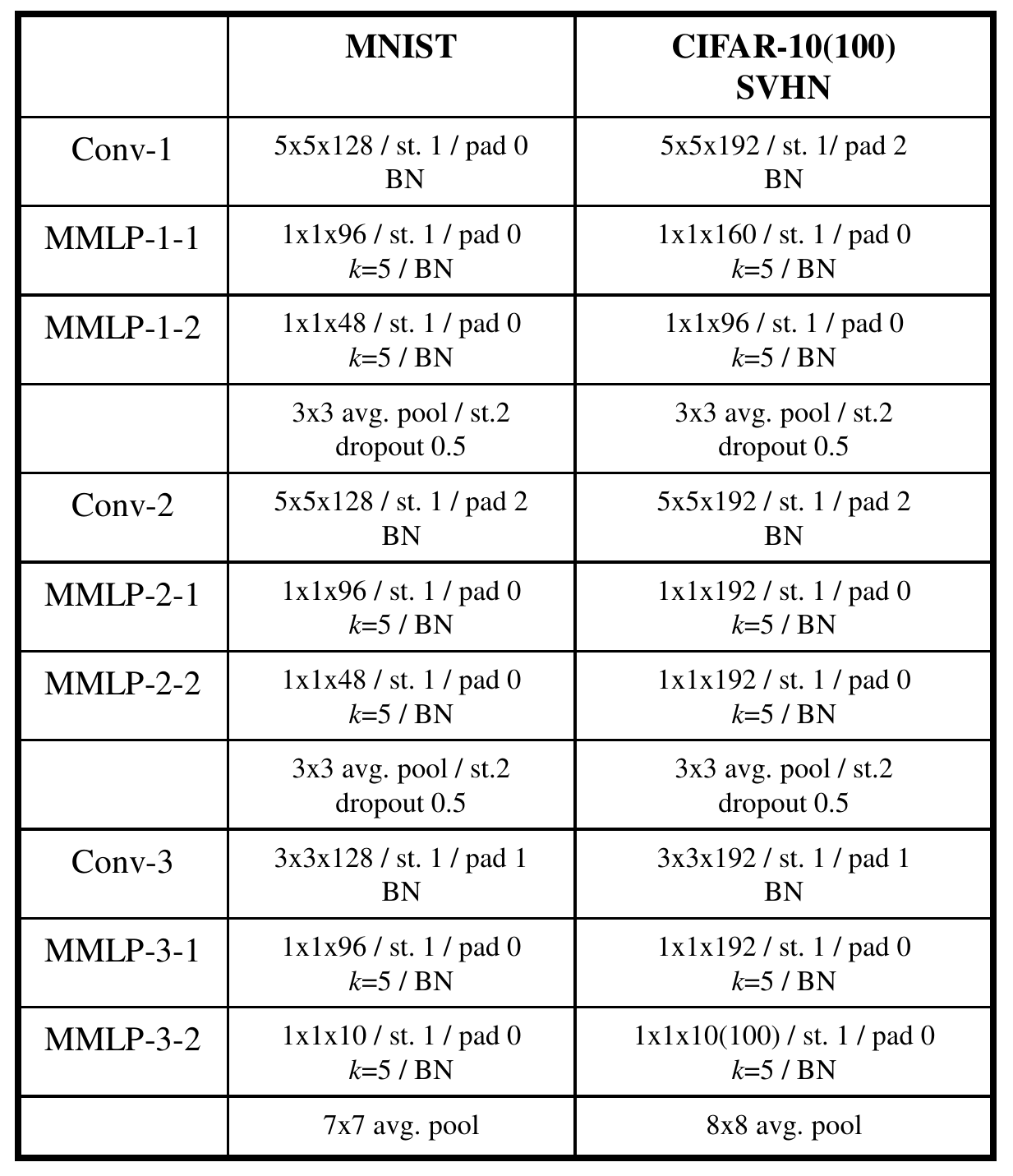}
\end{center}
\label{fig:para}
\end{table}

\subsection{MNIST}
The MNIST dataset~\cite{lecun1998gradient} consists of handwritten digit images, 28 x 28 pixels in size, organized into 10 classes (0 to 9) with 60,000 training and 10,000 test samples. 
Testing on this dataset was performed without data augmentation.
Table \ref{tab:MNIST} compares the results obtained in this study with those obtained in previous works.
Despite the fact that many methods can achieve very low error rates for MNIST dataset, we achieved a test error rate of 0.24\%, which set a new state-of-the-art performance without data augmentation. 

\begin{table}
\begin{center}
\caption{Comparison of test errors on MNIST without data augmentation, in which \textit{k} denotes the number of maxout pieces.}
\begin{tabular}{|p{1.6in}|p{0.6in}|} \hline 
Method & Error (\%) \\ \hline 
Stochastic pooling \cite{zeiler2013stochastic} & 0.47 \\ \hline 
Maxout network (\textit{k}=2) \cite{goodfellow2013maxout} & 0.47 \\ \hline 
NIN \cite{DBLP:journals/corr/LinCY13} & 0.45 \\ \hline 
DSN  \cite{lee2014deeply} & 0.39 \\ \hline 
MIM (\textit{k}=2) \cite{liao2015importance} & 0.35$\pm $0.03 \\ \hline 
RCNN-96~\cite{Liang_2015_CVPR} & 0.31 \\ \hline
\textbf{MIN} (\textit{k}=5)\textbf{} & \textbf{0.24} \\ \hline
\end{tabular}
\label{tab:MNIST}
\end{center}
\end{table}

%------------------------------------------------

\subsection{CIFAR-10}
The CIFAR-10 dataset~\cite{krizhevsky2009learning} consists of color natural images, 32 x 32 pixels in size, from 10 classes with 50,000 training and 10,000 test images. For this dataset, we applied global contrast normalization and whitening in accordance with the methods outlined in~\cite{goodfellow2013maxout}. To enable a comparison with previous works, the dataset was augmented by zero-padding 2 pixels on each side, which resulted in images 36 x 36 pixels in size. We then performed random corner cropping back to 32 x 32 pixels as well as random flipping on the fly during training.
Table \ref{tab:CIFAR10} compares our results with those obtained in previous works.
We obtained an error rate of 7.85\% without data augmentation and 6.75\% with data augmentation. These are the best results achieved to our knowledge. 

\begin{table}
\begin{center}
\caption{Comparison of test errors on CIFAR-10 dataset}
\begin{tabular}{|p{1.6in}|p{0.6in}|} \hline 
Method & Error (\%) \\ \hline 
\multicolumn{2}{|p{1.6in}|}{\textit{No data augmentation}} \\ \hline 
Stochastic pooling~\cite{zeiler2013stochastic} & 15.13 \\ \hline 
Maxout network (\textit{k}=2)~\cite{goodfellow2013maxout} & 11.68 \\ \hline 
NIN~\cite{DBLP:journals/corr/LinCY13} & 10.41 \\ \hline 
DSN~\cite{lee2014deeply} & 9.69 \\ \hline 
RCNN-160~\cite{Liang_2015_CVPR} & 8.69 \\ \hline
MIM (\textit{k}=2)~\cite{liao2015importance} & 8.52$\pm $0.20 \\ \hline 
\textbf{MIN} (\textit{k}=5)\textbf{} & \textbf{7.85} \\ \hline
\multicolumn{2}{|p{1.6in}|}{\textit{Data augmentation}} \\ \hline 
Maxout network (\textit{k}=2)~\cite{goodfellow2013maxout} & 9.38 \\ \hline 
NIN~\cite{DBLP:journals/corr/LinCY13} & 8.81 \\ \hline 
DSN~\cite{lee2014deeply} & 8.22 \\ \hline 
RCNN-160~\cite{Liang_2015_CVPR} & 7.09 \\ \hline
\textbf{MIN} (\textit{k}=5)\textbf{} & \textbf{6.75} \\ \hline
\end{tabular}
\label{tab:CIFAR10}
\end{center}
\end{table}

%-------------------------------------------
\subsection{CIFAR-100}
The CIFAR-100 dataset~\cite{krizhevsky2009learning} is the same size and format as the CIFAR-10; however, it contains 100 classes. Thus, the number of images in each class is only one tenth of that of CIFAR-10. As a result, this dataset is far more challenging. We applied the hyper-parameters used for CIFAR-10, but re-tuned the learning rate decay. This resulted in an error rate of 28.86\% without data augmentation, which represents the state-of-the-art performance. Table \ref{tab:CIFAR100} presents a summary of the best results obtained in previous works and the current work.

\begin{table}
\begin{center}
\caption{Comparison of test errors on CIFAR-100 dataset}
\begin{tabular}{|p{1.6in}|p{0.6in}|} \hline 
Method & Error (\%) \\ \hline 
Stochastic pooling~\cite{zeiler2013stochastic} & 42.51 \\ \hline 
Maxout network (\textit{k}=2)~\cite{goodfellow2013maxout} & 38.57 \\ \hline 
NIN~\cite{DBLP:journals/corr/LinCY13} & 35.68 \\ \hline 
DSN~\cite{lee2014deeply} & 34.57 \\ \hline 
RCNN-160~\cite{Liang_2015_CVPR} & 31.75 \\ \hline
MIM (\textit{k}=2)~\cite{liao2015importance} & 29.20$\pm $0.2 \\ \hline 
\textbf{MIN} (\textit{k}=5) & \textbf{28.86} \\ \hline
\end{tabular}
\label{tab:CIFAR100}
\end{center}
\end{table}

%-------------------------------------------------------------------------
\subsection{SVHN}
 The SVHN dataset consists of color images of house numbers (32 x 32 pixels) collected by Google Street View. There are 73,257 and 531,131 digits in the training and additional sets, respectively. In accordance with previous works~\cite{goodfellow2013maxout}, we selected 400 samples per class from the training set and 200 samples per class from the additional set for validation. The remaining 598,388 images were used for training. Moreover, there are 26,032 digits in the test set. For SVHN dataset, we applied the hyper-parameters as those used in the experiments mentioned previously, but re-tuned the learning rate decay. We also preprocessed the dataset using local contrast normalization, in accordance with the method outlined by Goodfellow et al.~\cite{goodfellow2013maxout}. Without data augmentation, we achieved a test error rate of 1.81\%, which is comparable to the best result obtained in previous works. Table \ref{tab:svhn} presents a comparison of our test results with those obtained in recent studies.

\begin{table}
\begin{center}
\caption{Comparison of test errors on SVHN. Note that Dropconnet~\cite{wan2013regularization} uses data augmentation and multiple model voting}
\begin{tabular}{|p{1.6in}|p{0.6in}|} \hline 
Method & Error (\%) \\ \hline 
Maxout network (\textit{k}=2)~\cite{goodfellow2013maxout} & 2.47 \\ \hline 
NIN~\cite{DBLP:journals/corr/LinCY13} & 2.35 \\ \hline 
Human performance~\cite{sermanet2012convolutional} & 2.00 \\ \hline 
MIM (\textit{k}=2)~\cite{liao2015importance} & 1.97$\pm $0.08 \\ \hline 
Dropconnect~\cite{wan2013regularization} & 1.94 \\ \hline 
DSN~\cite{lee2014deeply} & 1.92 \\ \hline 
MIN (\textit{k}=5)\textbf{} & 1.81 \\ \hline 
\textbf{RCNN-192}~\cite{Liang_2015_CVPR} & \textbf{1.77} \\ \hline
\end{tabular}
\label{tab:svhn}
\end{center}
\end{table}

%--------------------------------------------------------------------------
\subsection{Model capacity}

\begin{figure}
\begin{center}
	\includegraphics[width=3.15in, height=2.7in, keepaspectratio=false]{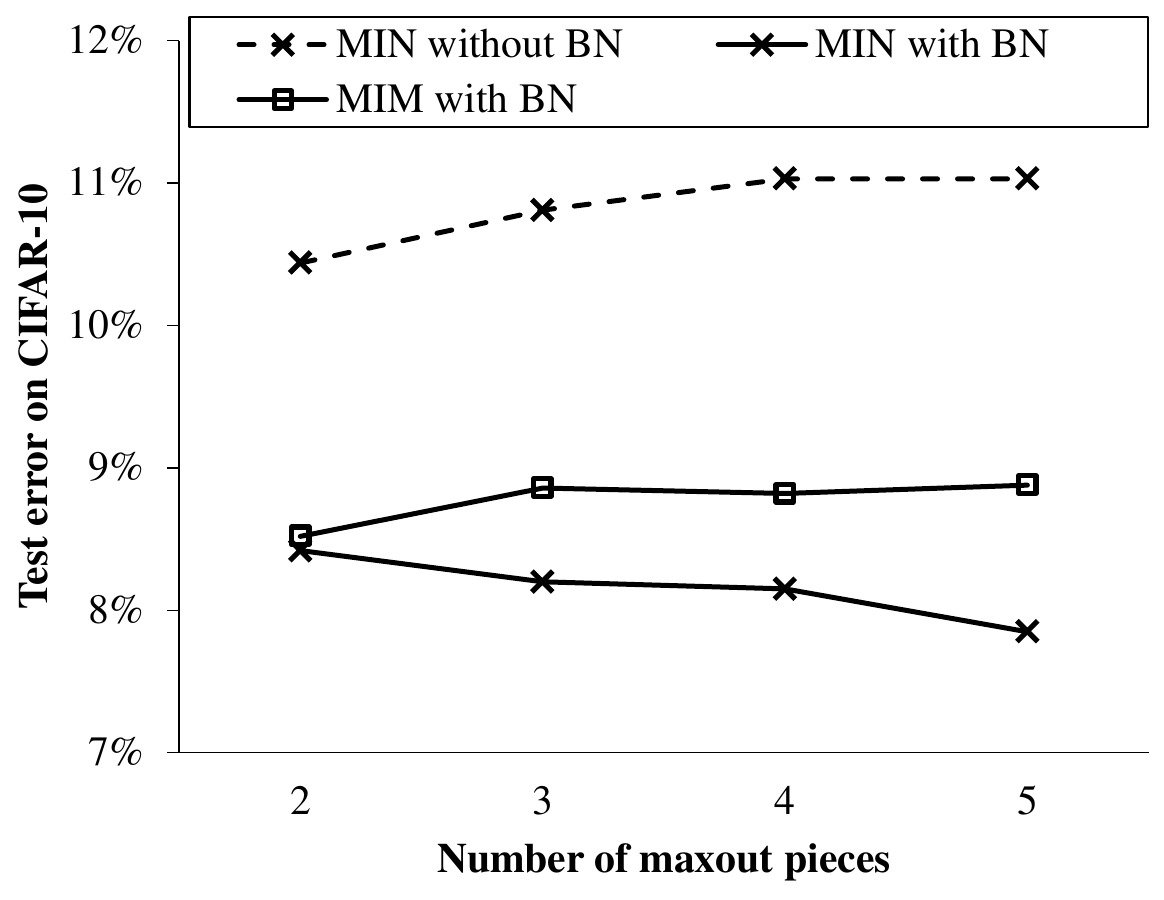}
\end{center}
   \caption{Performance related to the number of maxout pieces. We fixed the hyper-parameters when training the MIN model with different maxout pieces. Our method dramatically reduces test error of CIFAR-10 dataset with increasing the number of maxout pieces.}
\label{fig:comfig}
\end{figure}

 According to Montufar et al.~\cite{montufar2014number}, a deep neural network using ReLU activation function with $n_0$ input and $L$ hidden layers of width $n\ge $ $n_0$ can have $\Omega \left({\left({n}/{n_0}\right)}^{\left(L-1\right)n_0}n^{n_0}\right)$ linear regions. A deep maxout network with $L$ layers of width $n$ and $k$ maxout pieces can compute functions in at least $k^{L-1}k^n$ linear regions. This theoretical result indicates that the number of linear regions in a maxout network grows with the number of maxout pieces. Moreover, the number of linear regions in both ReLU and maxout networks grows exponentially with the number of layers. From this perspective, the Maxout network In Maxout network (MIM) model~\cite{liao2015importance} is indeed more complex than the proposed method.
However, the maxout network is prone to overfitting to the training dataset without model regularization. This can be attributed to the fact that the maxout network identifies the most valuable representations in the input during training, and it is prone to feature co-adaption. Therefore, MIM model in which three maxout layers are stacked in one MIM block may lead to overfitting and increasing the number of maxout pieces may not improve performance. We tested the proposed method on CIFAR-10 dataset using various numbers of maxout pieces. Figure \ref{fig:comfig} illustrates how increasing the number of maxout pieces can improve the performance of our method, by which point the MIM model has already reached saturation. This figure also shows the saturation of maxout units due to the growing number of maxout pieces without batch normalization. 

%------------------------------------------------------------------------
\subsection{Regularization of average pooling in MIN}
	Most of the previous methods used max pooling for down sampling. Max pooling extracts the features within local patches that are the most representative of the class.  In this study, the MIN block is able to abstract representative information from every local patch such that more discriminable information is embedded in the feature map. Thus, we are able to use spatial average pooling in each pooling layer to aggregate local spatial information. We then compared the results using average pooling in the first two pooling layers with those using max pooling, whereas the last pooling layer was fixed to global average pooling. Table~\ref{tab:avg} presents the test error associated with different pooling methods. The use of average pooling in all pooling layers was shown particularly effective. The irrelevant information in the input image can be inhibited by average pooling, but may be discarded by max pooling.  Average pooling is an extension of global average pooling in which the model seeks to extract information from every local patch to facilitate abstraction to the feature maps.

\begin{table}
\begin{center}
\caption{Comparison of test errors on CIFAR-10 dataset without data augmentation using max/average pooling in the first two pooling layers}
\begin{tabular}{|p{1.8in}|p{0.8in}|} \hline 
Method & Test error (\%) \\ \hline 
MIN (k=5)  + max pooling & 8.78 \\ \hline 
MIN (k=5)  + avg pooling & 7.85 \\ \hline 
\end{tabular}
\label{tab:avg}
\end{center}
\end{table}

\begin{figure*}
\begin{center}
	\includegraphics*[width=6.8in, height=4.65in, keepaspectratio=false]{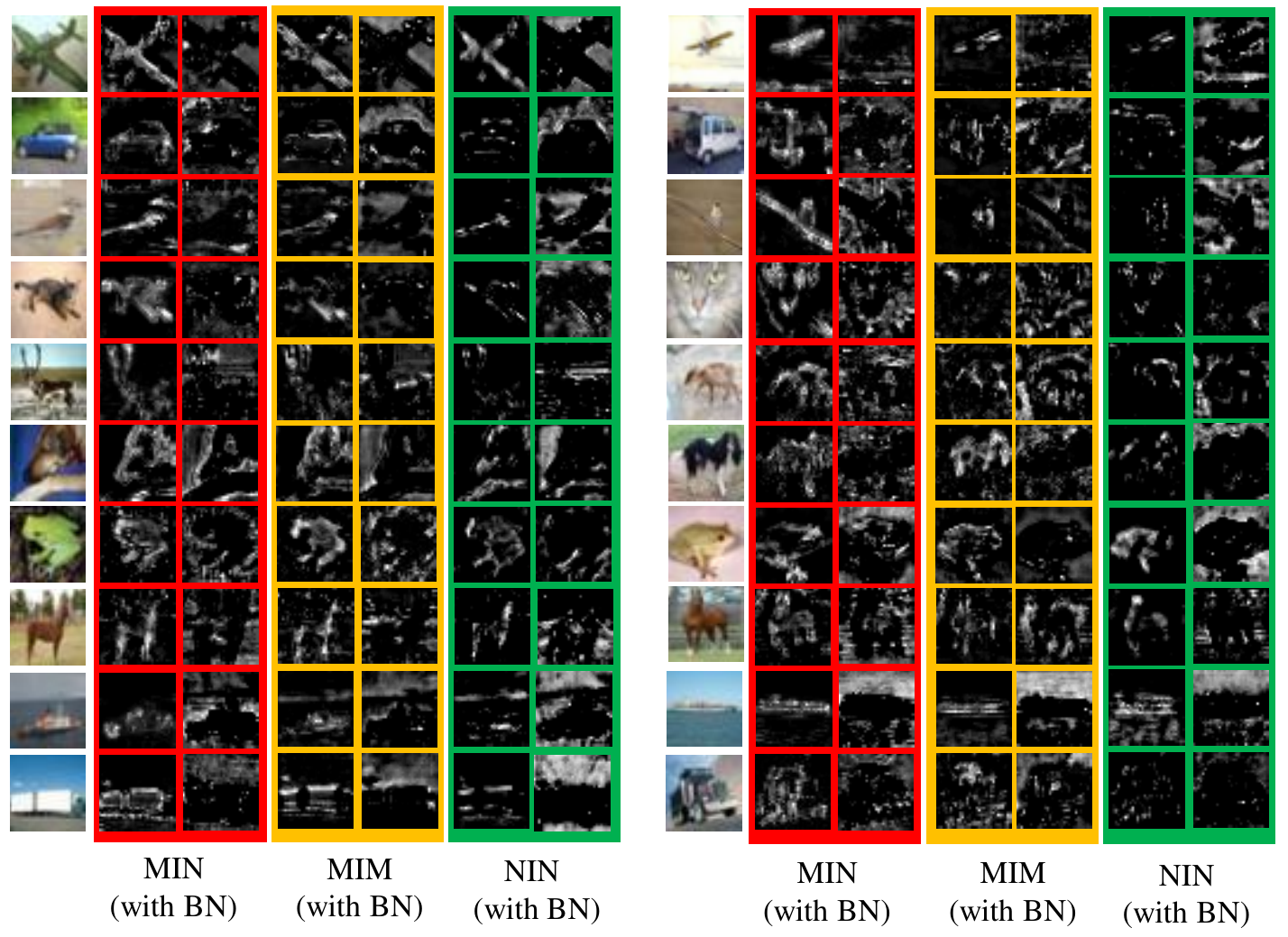}
\end{center}
   \caption{Visualization of learned feature maps before the first pooling layer obtained using the MIN, MIM, and NIN methods. Only the top 50\% of the data in each channel are presented.}
\label{fig:cifar}
\end{figure*}

\begin{figure*}
\begin{center}
	\includegraphics*[width=6.6in, height=3.5in, keepaspectratio=false]{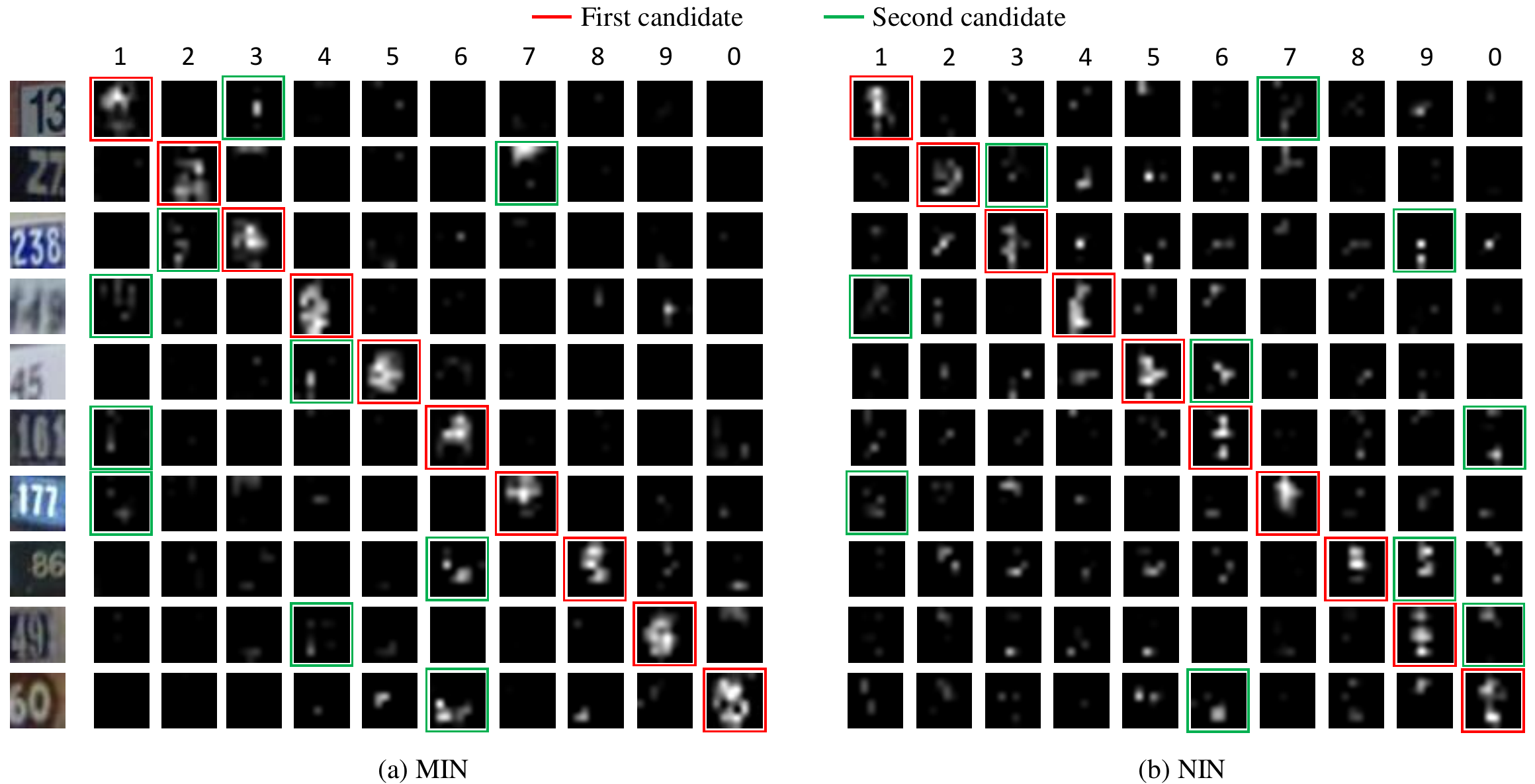}
\end{center}
   \caption{Visualization of the learned feature maps before the global average pooling layer obtained using the MIN and NIN methods. Only the top 10\% of the data are presented. The first and second candidates of the output are highlighted in red and green boxes. These results demonstrate the possibility applying the proposed MIN method to multiple object recognition.}
\label{fig:svhn}
\end{figure*}

%-----------------------------------------------------------------------
\subsection{Visualization of learned features}
 Average pooling was applied in all pooling layers to facilitate the abstraction of input images. We extracted feature maps from models trained using CIFAR-10 to illustrate these effects. Figure \ref{fig:cifar} presents examplar images and their corresponding feature maps, which were selected from the CIFAR-10 test set. For each method, the first column illustrates the selected feature maps related to the objects per se, whereas the second column shows the selected feature maps for the background.  Note that only the top 50\% of the data in each channel are shown in this figure. The learned feature maps produced using the proposed MIN method appear to be more intuitive than the other methods when dealing with both foreground and background. This finding demonstrates the effectiveness of MIN and its potential for object segmentation.

 Figure~\ref{fig:svhn} presents examplar images selected from SVHN test set and their corresponding feature maps extracted in the last convolutional layer by using the MIN and NIN models. Only the top 10\% of the data are presented. One of the major difficulties in the classification on SVHN dataset is that there are a large portion of images containing distractors, which may confuse the model during training. After all, the distractors are also digits and should  be recognized by the model during testing as well as the targeted digit in the center. Therefore, the model should recognize the distractors as the runners-up, besides classifying the targeted digit as the first candidate. In Figure~\ref{fig:svhn}, we presented the images containing targeted digits from 0 to 9 and distractors on the side and highlighted the first and second candidates of the output determined by the softmax layer. These results show that the proposed approach is able to recognize distractors in input images with high accuracy. 
 This indicates that the MIN can robustly preserve information of each category because of the pathway encoding in maxout MLP and spatial average pooling. When convolutional filters slide onto the distractor, the MIN model can extract features of the distractor along its own pathway. Moreover, the MIN model downscale the feature maps by using spatial average pooling and this pooling method keeps all information of a local patch, whereas max pooling only passes the maximal part. These results suggest the possibility of applying the MIN method to multiple object recognition using a more comprehensive image dataset, such as ImageNet.

In human visual system, the competing process of distractors has been investigated by the Eriksen flanker task~\cite{eriksen1974effects}. In this task, subjects are instructed to respond with one hand if the presented central letter is `H' and with the other hand if the letter is `S'. In general, subjects respond faster and more accurately when the center and flankers are the same (\eg{, HHHHH or SSSSS}) than when they are different (\eg{, HHSHH or SSHSS}).
Psychophysiological analysis~\cite{gratton1988pre} supported the theory indicating that the flankers activate the incorrect response competing with the correct response~\cite{eriksen1979information}.
That is, visual system processes all of the objects in the visual field and inhibits responses of incongruent objects. 
This suggests that the proposed MIN model resembles the human visual system such that the distractors are recognized and inhibited as runner-ups.

%-----------------------------
\section{Conclusions}
 This paper presents a novel deep architecture, MIN. A MIN block, consisting of a convolutional layer and a two-layer maxout MLP, is used to convolve the input and average pooling is applied in all pooling layers. In neuroscience perspective, the proposed architecture is similar to the mechanism of visual system in the brain. 
The proposed method outperforms others because of the following improvements: the MIN block facilitates the information abstraction of local patches, batch normalization prevents covariate shift, and average pooling acts as a spatial regularizer tolerating changes of object positions. Our experiments showed that the MIN method achieves state-of-the-art or comparable performance on the MNIST, CIFAR-10, CIFAR-100, and SVHN datasets. Moreover, the extracted feature maps demonstrate the efficacy of categorical representation by using the MIN method, as well as its potential to multiple object recognition.

%-------------------------------------------------------------------------

{\small
\bibliographystyle{ieee}
\bibliography{egbib}
}

\end{document}